# A Semi-supervised Approach for Activity Recognition from Indoor Trajectory Data

Mashud Rana, Ashfaqur Rahman, and Daniel Smith

*Abstract*—The increasingly wide usage of location aware sensors has made it possible to collect large volume of trajectory data in diverse application domains. Machine learning allows to study the activities or behaviours of moving objects (e.g., people, vehicles, robot) using such trajectory data with rich spatiotemporal information to facilitate informed strategic and operational decision making. In this study, we consider the task of classifying the activities of moving objects from their noisy indoor trajectory data in a collaborative manufacturing environment. Activity recognition can help manufacturing companies to develop appropriate management policies, and optimise safety, productivity, and efficiency. We present a semi-supervised machine learning approach that first applies an information theoretic criterion to partition a long trajectory into a set of segments such that the object exhibits homogeneous behaviour within each segment. The segments are then labelled automatically based on a constrained hierarchical clustering method. Finally, a deep learning classification model based on convolutional neural networks is trained on trajectory segments and the generated pseudo labels. The proposed approach has been evaluated on a dataset containing indoor trajectories of multiple workers collected from a tricycle assembly workshop. The proposed approach is shown to achieve high classification accuracy (F-score varies between 0.81 to 0.95 for different trajectories) using only a small proportion of labelled trajectory segments.

*Index Terms*—Activity recognition, behaviour classification, trajectory analytics, smart manufacturing, deep learning, semi-supervised model.

## I. INTRODUCTION

MANUFACTURING systems form a vital part of the society and economy, providing jobs to workers and products to consumers. The manufacturing industry is continuously developing new initiatives (such as Industry 4.0, Industrial Internet) to transform traditional manufacturing paradigms. Smart manufacturing is a technology-driven approach to improve various factors affecting the performance of manufacturing systems through the integration of sensor data, analytics and automation [1]. The computational analysis of data streams collected from

This research was supported by the Future Digital Manufacturing Fund (FDMF) at Data61, Commonwealth Scientific and Industrial Research Organisation (CSIRO), Australia. Thanks to Reena Kapoor, Peter Baumgartner, and Elena Tartaglia for providing feedback on different sections of the paper.

Mashud Rana, Ashfaqur Rahman, and Daniel Smith are with Data61, CSIRO, Australia. Emails: mdmashud.rana@data61.csiro.au, ashfaqur.rahman@data61.csiro.au, daniel.v.smith@data61.csiro.au.

Internet of Things (IoT) based sensors at various stages of production enables evidence-based decision making, and provides a systematic way to monitor and improve manufacturing systems [2, 3].

Numerous studies investigated the diverse machine learning and data mining techniques for analysing heterogenous data for a great variety of manufacturing applications. A few prominent applications include fault detection and diagnostic [4], predictive analytics [5], quality control and monitoring [6], efficiency monitoring [7], process optimisation [8], product life cycle management [9], workflow evaluation [10], indoor space modelling [11], streamlining supply chains [12], and safety [13]. Most of these studies primarily focused on utilising the data collected from workstations or equipment via accelerometers, gyroscopes, magnetometers etc. Moreover, many of the key dynamics of a manufacturing process can be observed from trajectory data that has been collected by using tracking devices attached to entities or objects. Trajectory data consists of the spatial coordinates of an object as a function of time as well as the features describing the object that is being tracked. In a manufacturing system, the movement of materials, vehicles, products, logistics, tools, and workers can be used to determine its current state. Translating these observations into meaningful insights is the realm of *movement analytics* [14], where machine learning and other statistical methods are used to uncover object semantics based on its movement patterns.

In this study, we are interested in a specific aspect of movement analytics, viz., *activity recognition*. Activity recognition is generally defined as the task of identifying the actions of objects from a series of motion related measurements captured by sensors [15, 16]. It can play a vital role to identify inefficiencies or bottlenecks within manufacturing processes. It also helps to identify the possible reasons behind the deviation of the relevant entities from their expected behaviours. The decision makers can use this information to develop appropriate management policies, and optimise the safety, productivity, and efficiency. For example, a worker may move back and forth between workstations for several reasons including delivery of items produced to co-workers for subsequent processing, finding misplaced or missing tools. Obviously, it is a problematic activity and indicative of operational inefficiencies if the visits are related to searching for misplaced tools or components. Hence, identifying which tools or workstations are involved in these activities could help to identify the problems within the manufacturing process which is a prerequisite for improving operational efficiency.

Although activity recognition has widespread application across different fields (e.g., retail ([17, 18]), health [19, 20]), tourism [21], construction [22]), it has rarely been studied with respect to manufacturing applications. Several studies (e.g., [23-25]) provide insights on the current state of the literature to detect, recognize, and monitor activities utilising diverse datasets and methods. The two main approaches for activity recognition are either *vision-based* or *sensor-based* [24]. Vision-based approaches utilise cameras to passively capture video of the entities of interest, while sensor-based approaches commonly use time series from motion sensors (i.e., accelerometers, gyroscopes) attached to the entities. In contrast to most of these previous studies, we aim to develop a model for human activity recognition using spatiotemporal trajectories acquired with sensor-based localisation technologies. Activity recognition especially from indoor trajectory data is challenging due to spontaneous nature of human movement within a limited indoor space [16]. In contrast to outdoor trajectory data of the objects moving over large geographical area (e.g., vehicle or tourist trajectories in cities, vessels trajectories in oceans), indoor trajectory covers a small area with overlaps. The frequent movement of human within a small indoor area makes the indoor trajectory data noisy and difficult to model. Moreover, majority of activity recognition approaches use supervised machine learning models, which rely upon large sets of labelled data (ground truth) for the training of classification models [15]. Generating labelled datasets is a labour intensive and costly process that can be a major bottleneck for using supervised machine learning. For activity recognition, it is often unwieldly to manually label the large number of segments pertaining to the different activities that may be present across time. Hence, in this study we adopt semi-supervised learning, which only requires fewer labelled segments as examples to train the activity recognition model.

Specifically, our contributions in this paper can be summarised as follows. We develop an approach for human activity recognition from spatiotemporal trajectory data in an indoor environment. The proposed approach consists of three steps. Firstly, each complete trajectory is partitioned into a set of segments by optimising an information coding metric, the minimum description length. Partitioning allows to identify and classify the different activities of a worker across different time periods of their work shift. Secondly, the pseudo labels of the trajectory segments were then generated based on a constrained hierarchical clustering which requires a small set of labelled segments. Finally, a Convolutional Neural Network (CNN) based deep learning model is trained using the pseudo labels of the trajectory segments. The model was then used to classify the raw trajectories into a sequence of worker activities. While the different components of the developed approach (like trajectory partitioning, clustering, convolutional neural networks) are all well-known, we emphasise that the overall architecture combining these components is new for semi-supervised activity recognition especially utilising indoor trajectory data. We evaluate the proposed approach by using a dataset of workers' trajectories that were collected with a motion capture positioning system during a tricycle assembly process. We aim to identify a set of target activities specific to the manufacturing working environment: *standing at workstation*, *moving randomly*, *moving between workstations*, *restocking*. However, we note that our approach is generic and can be applied to such other activities as well, thus preserving the motivation for helping improve operational efficiency. This study is the first of its kind to design and develop a semi-supervised machine learning activity recognition model based on indoor trajectory data for manufacturing application.

## II. Related Work

The technological innovations in tracking systems and IoT based sensors have made it possible to collect data from moving objects over space and time. In the era of cyber-physical systems, machine learning has been an integral part of smart manufacturing to develop analytics utilising such data for intelligent decision making [5]. Numerous studies (e.g., [6, 7, 26-28]) in the literature investigated the application of machine learning to support different applications. In this section, we review previous research related to the application of machine learning in manufacturing. Whilst reviewing the literature, we limit the scope to movement analytics in the manufacturing domain and activity recognition in general.

### A. Movement Analytics in Manufacturing

Movement analytics (also known as trajectory data analytics) refers to the process of extraction and utilisation of knowledge from tracking data for providing meaningful solutions to decision makers [14]. The spatiotemporal tracking data can be utilised for optimising the production processes in manufacturing and developing domain specific applications. Szabo et al. [10] studied the feasibility of different Real Time Locating Systems (RTLS) for capturing location data of moving objects to support different applications in manufacturing. They also presented a use case to identify the bottlenecks in defined production zones and measure cycle time deviation at the workstations in an automotive company. To identify the bottlenecks (in terms of temporary storage or unplanned workstations in the production process), they grouped the position data by applying the k-means clustering algorithm. The cycle time of workstations was measured based on classified zone data that were visualized later to provide real-time status of the production process. Arkan and Van Landeghem [29] considered improving Work-in-Process (WIP) visibility in the semi-automated shop floor of an automotive manufacturing company. They utilised the spatiotemporal data collected with RTLS from a multi-item production line to compute a set of Key Performance Indicators (KPIs) such as cycle time, cycle speed, production time, defect reject ratio and workspace utilization. These KPIs were then analysed to evaluate the existing workflow and redesign the floor with a simulation tool. Similarly, Gyulai et al. [30] developed analytics to compute KPIs (from simulated trajectory data) to evaluate the performance of a production system and facilitate the implementation of situation aware production control.

Moreover, different objects are likely to work together in a collaborative manufacturing environment. Therefore, it is important for the objects to efficiently and accurately identify the task plans of others and respond in a safe manner. Chen et al. [13] presented an analytics framework to predict human trajectories and to infer the work plan to facilitate safe and effective collaboration between objects (human and robots). A Long Short-Term Memory (LSTM) recurrent networks was used to model the temporal dynamics of sequential movement data and Bayesian inference method was applied to infer the potential plans of workers utilising LSTM based predictions. Locklin et al. [31] also predicted future positions of human trajectories by fitting a second degree polynomial function to historical location data to enable collaboration amongst a large number of workers in the indoor space. Zhang et al. [32] proposed an LSTM based method to predict the future motion trajectory of human operators for facilitating a robot's action planning and execution in a car engine assembly factory. Wang et al. [33] proposed a similarity based model for trajectory prediction within indoor spaces. Specifically, they applied the k-Nearest Neighbours (kNN) algorithm to find a trajectory from the database that was most similar to the given trajectory. The next location of the given trajectory was then predicted from the path of similar trajectory. The main novelty of this model was the formulation of a distance metric that considered both the spatial and semantic distance between trajectories, which were computed based on the longest common sub sequences and dynamic time warping, respectively.

Furthermore, the optimal utilisation of available indoor space (such as shop floor, production floor, etc.) is vital for mass production. Movement data can be used to understand the geospatial interaction patterns between objects, and hence, helps to design more efficient factory layouts. Han et al. [11] presented a method to study indoor space utilisation based on the common movement patterns in the trajectory data of multiple users. They first partitioned each trajectory into a set of segments by optimising an information theoretic criterion and then grouped the segments from all of the trajectories into a set of clusters by applying the GDBSCAN algorithm [34]. The clustering results were used to identify heavily utilised regions and visualize the evolution of utilisation over time for the better design of indoor spaces in the future. Additionally, Cai et al. [35] proposed a spatiotemporal data model for monitoring IoT enabled production systems. Their model combined the principle of the Apriori algorithm [36] and depth first search to find the frequent trajectory patterns of WIP. Bu [37] also described a framework based on the Apriori algorithm to mine frequent path patterns from the massive amounts of tracking data. This enabled material flow paths to be adjusted and helped to reschedule the route of automatic guided vehicle robots in a production environment.

*B. Activity Recognition*

Recognising the activities of objects (people, vehicles, robots) is a key factor in developing appropriate strategies for industrial applications in many domains including but not limited to retail ([17, 18]), health ([19, 20]), construction ([22]) and transport ([38]). Shum et al. [39] reviewed utilisation of different types of tracking data collected in variety of domains for human activity recognition. Arslan et al. [22] developed a model for workers' activity recognition at hazardous construction sites for improving safety management strategies. The raw GPS data was transformed into semantic trajectories to label mobility related activities in terms of their building environment. A Hidden Markov Model (HMM) was then trained using the semantic trajectories to classify the mobility patterns of workers. Polanti et al. [17] developed an intelligent system to improve the shopping experience by utilising the movement trajectories of customers in retail environments. A HMM was trained with the shopping trajectories of customers in order to predict the customer's future shopping preferences. The system then presented a route map to direct customers to their preferred products in the retail store. Alahi et al. [38] proposed social LSTM, a model to predict human movement within crowds utilising their spatial trajectories. Given the motion of individuals within a crowded space are affected by the behaviour of others, the social LSTM architecture modelled these spatial interactions. The movement of individuals were represented by separate LSTMs and a shared pooling layer was used to connect the latent states of all the individuals (their LSTMs) within the crowded space. The social LSTM was then used to predict the future movement of individuals and groups of individuals.

Several studies investigated activity recognition from indoor tracking data to detect early symptoms of abnormal behaviours or health risks. Gochoo et al. [19] presented a non-obtrusive activity recognition model for elderly people living alone. The indoor tracking data of individuals was converted into two-dimensional activity images that were then used to train a Deep Convolutional Neural Network (DCNN) with a predefined set of home activity classes. The application of DCNN and other machine learning models including Random Forest (RF) and Gradient Boosting (GBM) were also investigated in [40] for the detection of dementia related behaviours of elderly people using movement data. Similarly, Fang et al. [41] identified abnormal behaviour patterns linked with different health risks in order to prevent their occurrence. A hybrid model based on an LSTM and Grey Model was proposed using the past movement data of an individual to predict their future location and activity class. In contrast to the continuous valued position data considered in our study, the trajectory data used in [19, 40, 41] consists of binary on/off signals from a set of fixed indoor sensors.

Moreover, Yu et al. [42] presented a feature-oriented method for identifying truck parking behaviours from the vehicle's trajectory data. The raw GPS trajectories were processed to extract a set of exploratory features and then association rule mining was applied to the extracted features to identify legal and illegal parking patterns. Lei [43] described a framework to identify the anomalous behaviour of vessels travelling in maritime space from their trajectories. The framework mapped the trajectories into spatial regions by applying a grid based clustering algorithm and then extracted features reflecting the spatial, sequential, and behavioural characteristics. The movement behaviours of the vessels were then classified using a probabilistic suffix tree which utilised those features as inputs. Likewise, Chatzikokolakis et al. [44] applied RF model on vessels trajectories to identify search and rescue activities. Kim et al. [45] developed a clustering based method to discover travel patterns utilising vehicle trajectory data in a traffic network. The vehicle trajectories were first grouped by applying a density based clustering algorithm using the Longest Common Subsequence (LCS) distance metric. The overlapping LCS from all the clusters were then merged using hierarchical clustering to generate travel patterns representative of the dataset. New trajectories were then classified by matching them against the clusters of representative travel patterns. Song et al. [46] presented a deep recurrent neural network architecture to simultaneously solve multiple learning tasks using spatial trajectories across large scale transportation networks. The future motion of an individual and their transportation mode were simultaneously predicted using a hierarchical network of LSTMs that represent motion across different temporal scales. Two LSTM based encoders were utilised to represent the inputs of each task separately, two LSTMs were used to create a shared feature representation and a pair of LSTM decoders that were used to generate the outputs of each task.

*C. Summary*

Trajectory data can play a vital role in smart and adaptive manufacturing. However, the above review indicates recent activity recognition methods utilising tracking data have not been well studied in the context of manufacturing. The existing studies primarily utilised trajectory data for workflow evaluation. There exists many applications that required regular monitoring of activities or understanding the behaviours of moving objects [47]. A few examples of such applications include: i) monitoring suspicious activities in large industrial workshops or chemical plants for security purposes, ii) identifying when a worker interacted with other workers to avoid spreading infectious diseases (e.g., COVID) and loss of workforce, iii) understanding why and to what extent a worker deviated from their intended workflow. On the other hand, while numerous research investigated utilisation of outdoor trajectory data (collected using GPS) for different applications, activity recognition based on indoor tracking data is not sufficiently studied especially for manufacturing applications. Indoor tracking data covers small area with overlaps. Contrast to outdoor trajectories that are relatively smoothed, the segments of indoor trajectories are significantly shorter and noisy that make the activity recognition task very difficult. Given most of the existing methods for activity recognition are supervised, they are not well suited to manufacturing applications due to the difficulties in collecting labelled activity data in such contexts. Hence, it is also important to explore the feasibility of semi-supervised methods on indoor tracking data. In this study, we aim to address these deficits in the literature.

## III. DATASET

The indoor tracking dataset used in study was collected from a tricycle assembly workshop [48]. The workshop consists of multiple workstations and provides a dynamic environment for the workers during the assembly process with various representative industrial scenarios. The dataset is publicly available at [49].

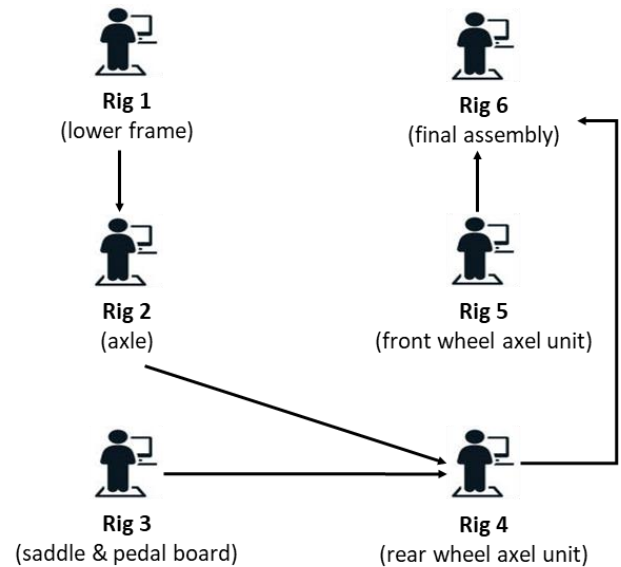

**Fig. 1.** Expected workflow of the people at the tricycle assembly workshop which consists of several workstations.

Fig. 1 shows the arrangement of the workstations of the tricycle assembly line and the expected movement scenarios of the people across the workstations during the assembly process. There are six workstations (also called rigs) in the assembly line. The worker at each workstation is responsible for building certain parts of a tricycle and concurrently works in a collaborative manner with others during the assembly process. The worker at rig 1 prepares the lower frame of tricycles and delivers it to the worker at rig 2 who assembles the axle with the lower frame. The worker at rig 3 builds the saddle and pedal board, and then supplies these components to the worker at rig 4. The worker at rig 4 builds the rear wheel axle unit by assembling the units built by the workers at rigs 2 and 3. The worker at rig 5 assemble the front wheel axle unit. Finally, both the front wheel axle and rear wheel axle units are assembled to finalise the tricycle construction by the worker at rig 6.

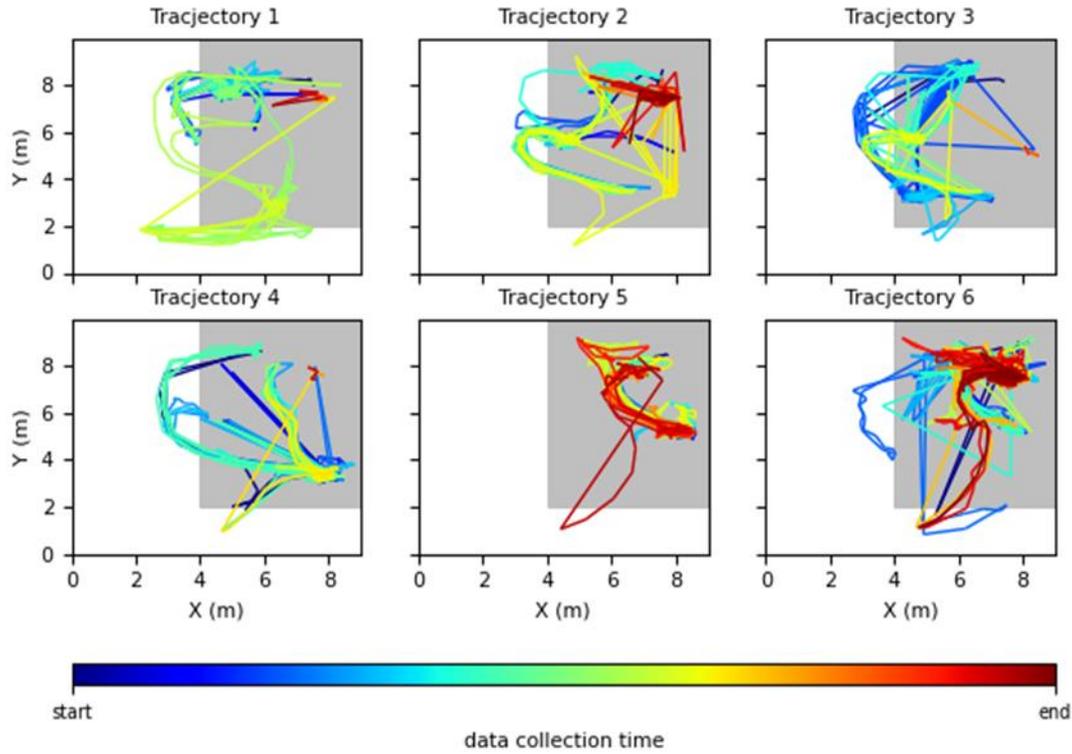

**Fig. 2.** The processed tracking data of the workers at the 10×9 square meters tricycle assembly workshop. The shaded area (▪) indicate the work zone where the workstations are placed, and the colors of the trajectory lines indicate data collection time.

Six tricycles are expected to be built within three hours of operation. The workstations can hold the components required for the assembly of three tricycles. Hence, the workers need to restock the required components from the storage area when initial stocking is finished after the first round of work, or if there are any missing components or tools. During the assembly process each person is responsible for the pre-assigned task, can go to help co-workers or take breaks as necessary. For more details on the site setup and data collection procedure, we refer to the previous study [48].

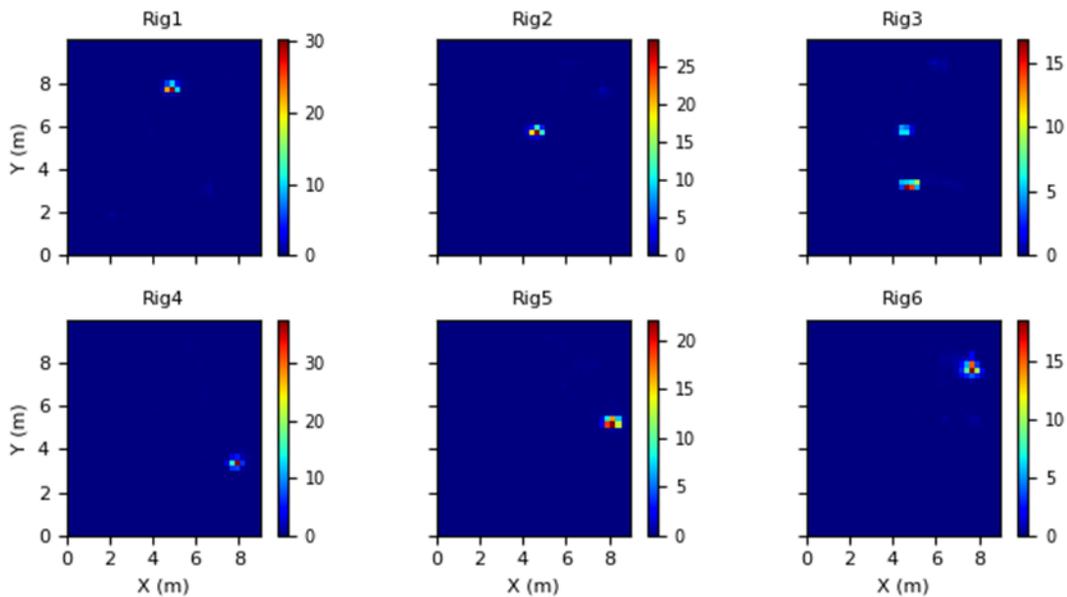

**Fig. 3.** Distribution (%) of the spatial points for six trajectories at the tricycle assembly workshop.

Movement data of the workers in the workshop were recorded using a Motion Captured (MoCap) system [50] for 3 hours of operation. Mocap systems provides better indoor positioning accuracy compared to other available technologies. The interval between consecutive data samples varies between 10 to 100 milliseconds. We down sample the raw trajectory data for each worker to 1 sample per second to synchronise the interval between successive samples and reduce the number of missing samples. **Fig. 2** shows the processed trajectory data of each worker during the entire data collection period. The tracking data shows that the workers deviate from their planned movement protocol and visit locations outside of the defined assembly zones as well. These deviations could be associated with taking a break or other unknown behaviours (e.g., activity of *moving randomly*).

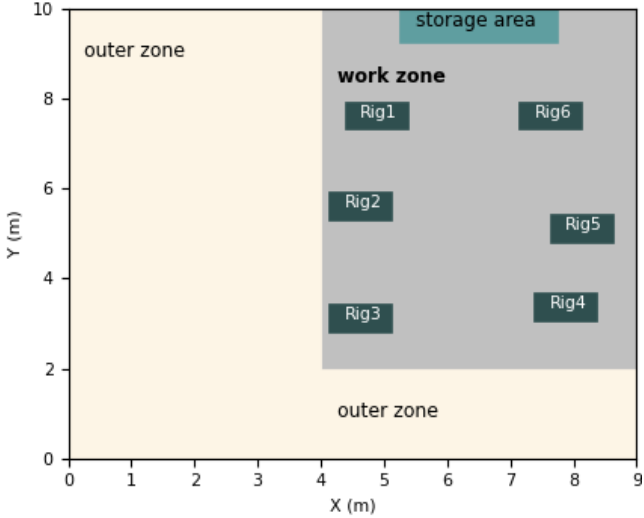

**Fig. 4.** Factory layout based on distribution of spatial points and site information.

Moreover, we analyse the spatial distribution of worker positions (**Fig. 3**) to identify the approximate location of the workstations and map them into a factory layout. The factory layout is required to manually label the different trajectory segments. For each worker, the location of the workstation is determined as the point of maximum density within the distribution of the worker's spatial positions over time, since each worker should spend majority of his/her time at the designated workstation to finish the assigned task. **Fig. 4** presents the approximate layout of the tricycle assembly line based upon the spatial distribution of the workers' trajectories. As it shows, the six rigs are placed to make the interactions or collaborations among the workers easy by following the expected movement scenarios shown in **Fig. 1**.

## IV. BACKGROUND

### A. Definitions

*Trajectory:* a *trajectory* of an object or entity is an ordered sequence of location points and is denoted as: $TR = \{(p_1, t_1), (p_2, t_2), \dots, (p_N, t_N)\}$ : $t_{i-1} < t_i$, $i = 2, \dots, N$, where $p_{i \, (1 \le i \le N)} \in R^d$ is a multidimensional vector of spatial location information of the object at timestamp $t_i$ and $N$ is the total number of location points in $TR$. In the simplest form, $p_{i \, (1 \le i \le N)} \in R^d$ represents the object's location in a two dimensional plane at timestamp $t_i$. However, the dimension of the vector $p_i$ can be extended further by adding more features (e.g., third spatial dimension, velocity, acceleration, etc.) depending on the application.

*Sub-trajectory:* a *sub-trajectory* or *trajectory segment* of a trajectory $TR$ is a subset of time ordered spatial location points in $TR$ and is denoted as: $SUB_{TR} = \{(p_k, t_k), (p_{k+1}, t_{k+1}), \dots, (p_{k+n}, t_{k+n})\}$ : $t_k < t_{k+1}, 1 \le k < n \le N - 1$. The straight line joining the two endpoints $(p_k, p_{k+n})$ of $SUB_{TR}$ is called a trajectory partition of $TR$.

### B. Problem Statement

Given a set of trajectories $\mathcal{T} = \{TR_1, TR_2, \dots, TR_W\}$, where each trajectory represents the movement patterns of an object in an indoor manufacturing environment. For each trajectory $TR_i$ in $\mathcal{T} = \{TR_i\}_{i=1}^{W}$, the goal is to partition the trajectory into a set of non-overlapping segments and then classify each segment as a category from a set of four predefined activities: [*moving randomly, standing at workstation, moving between workstations, restocking*].

The activity of an object is labelled as '*moving randomly*' for the duration of a trajectory segment if the object moves outside the work zone (i.e., outside of the tricycle assembly area as shown in **Fig. 4**) for a different purpose such as taking a break, meeting others, etc. '*Standing at workstation*' indicates that the object is busy at the designated workstation to finish the assigned tasks. '*Moving between workstations*' represents the collaborative behaviour – a worker may visit other workstations to deliver the products (or components) built according to the workflow or help co-workers. Finally, '*restocking*' refers to the activity of visiting storage area if the stock required to complete the assigned task at a workstation is finished, or when searching for tools or component if they are not available at the workstation as expected.

In a typical manufacturing setting, many objects (e.g., workers, AVG, robots, etc.) work collaboratively in a dynamic environment. Each of these objects generate its own trajectory over time. Segmenting each of the trajectories and labelling them manually are not feasible since it requires time, resources, and expertise. In this study, we aim to automatically segment the trajectories by identifying a set of changepoints (or characteristic points) and then develop a semi-supervised model that required only few labelled segments (can be as low as one labelled segment for each category of the activities or behaviours).

### C. Constrained Hierarchical Clustering

Clustering is the process of grouping the samples (or observations) in a dataset such that the samples (e.g., trajectory segments in our case) within the same group (called a cluster) are similar to one another and dissimilar to the samples in other groups. Agglomerative hierarchical clustering initially assigns each data sample into a separate cluster and then successively merges them using a bottom-up approach. In each iteration, the two closest or most similar pair of clusters are merged into a single cluster where the closeness or similarity is measured based on a linkage criterion. *Single link* refers to distance between the two nearest samples whereas *complete link* refers to the distance between the two farthest samples as the similarity between two clusters, respectively. *Average link* considers the average of the distances of each pair of samples in two clusters. The merging process is finished when all the data samples form a single cluster. This method produces a set of nested clusters in hierarchical structure that can be visualised using a dendrogram which is tree like diagram to record the sequence of merges. The expected number of clusters can be obtained by drawing lines at different levels on the dendrogram depending on the applications or specifying the number of clusters during the merging process.

Moreover, it is possible to specify the structural constraint into hierarchal clustering process in the form of *must link* and *cannot link*. *Must link* constraint indicates which group of data samples should be part of the same cluster whereas *cannot link* constraint refers to the samples that should be in different clusters. **Fig. 5** shows an example of agglomerative hierarchical clustering using structural constraints. The distance matrix indicates that data samples (*b*, *d*) should be part of same cluster (*must link*), whereas pairs of samples in (*a*, *e*) and (*g*, *f*) cannot be in same cluster (*cannot link*). The dendrogram for this example shows that there are two different clusters are possible at the top due to the cannot link constrained.

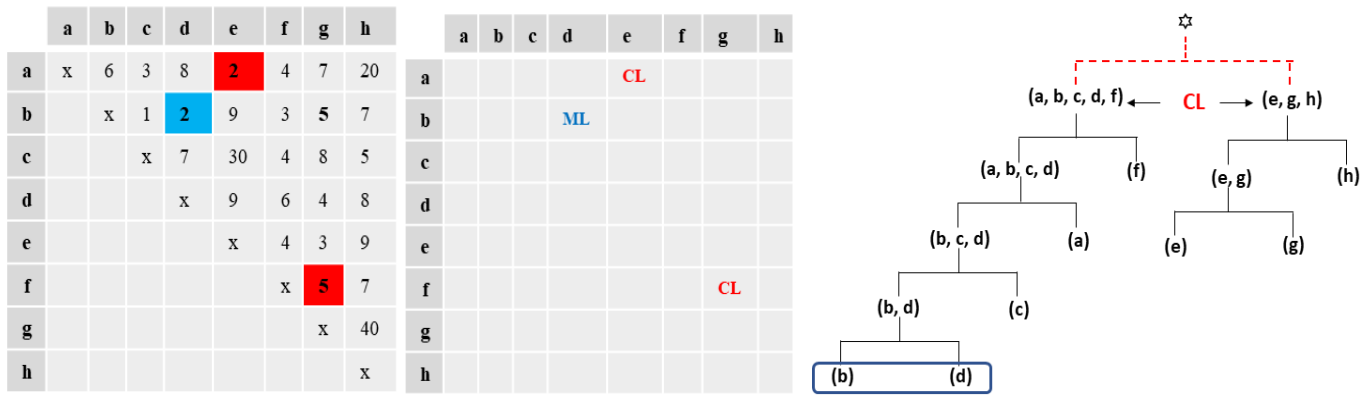

**Fig. 5.** Constraint agglomerative hierarchical clustering using both *must link* and *cannot link* constraint: distance measure between examples (left), '*must link*' and '*cannot link*' constraint (middle), and clusters merging process (right). In each iteration, pair of clusters are merged considering distance measured based on single link criteria.

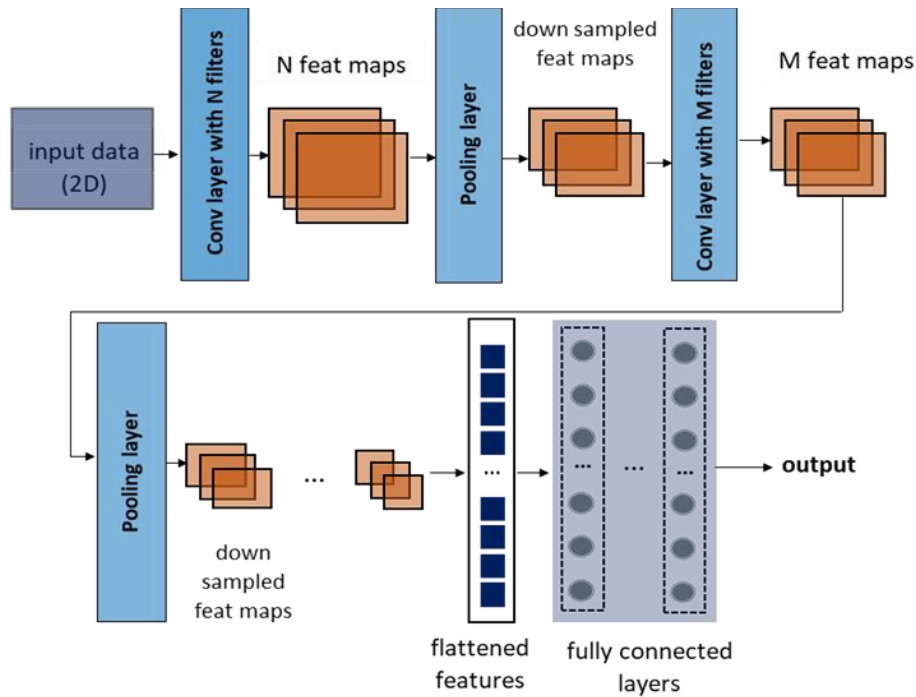

**Fig. 6.** A typical architecture of a convolutional neural networks

## D. Convolutional Neural Networks

CNNs [51] are prominent deep learning models that can identify the spatial patterns and translation invariant features from the input in a layered structure for classification or prediction. As shown in **Fig. 6**, a typical CNN architecture consists of a series of convolutional and pooling layers followed by one or more fully connected layers (also known as dense layers). The convolutional layers are the major building blocks of CNNs that apply a set of learnable filters (also known as kernels) to local regions of the input to extract useful features and create an internal network representation. The repeated application of the filters across different input locations creates a set of feature maps. The stacking of convolutional layers in a deep network allows the shallower layers to learn low-level features and the deeper layers to learn high-order or more abstract features.

The feature map outputs from the convolutional layers are location sensitive, that is, the layer outputs are dependent upon the feature's position within the data. To make the feature maps 'local translation invariance' (i.e., the output is not changed by local shifts in the feature position) and reduce the dimensionality of network representations, it is common for pooling layers to be added after the convolutional layers. Pooling is a down sampling operation that involves applying a sliding window to approximate local regions of the features. The approximation commonly involves computing the maximum or mean value within the feature window. As such, pooling has been considered as a technique to generalize feature representations. Overall, this structure allows the network to learn filters that represent patterns in the data that can be used for prediction or classification [52]. Pooling also makes the CNNs more noise tolerant and creates a hierarchy of features to extract meaningful patterns at different temporal scales [53].

The last part of a CNN is analogous to traditional feedforward NNs and consists of one or more dense layers. Before feeding the extracted feature maps to the fully connected layer, it is required that the feature maps are flattened into a vector. The dense layers are the final network layers that apply nonlinear combination of the extracted features to compute output predictions. For more detail information on CNNs, we refer to [51, 52, 54].

## V. PROPOSED APPROACH FOR ACTIVITY RECOGNITION

**Fig. 7** shows an illustrative diagram of our proposed approach for activity recognition from indoor trajectory data. The approach consists of three main steps: *trajectory partitioning*, *clustering*, and *model training*. The first step focuses upon partitioning each trajectory into a set of segments representing various movement patterns. This is achieved with a segmentation algorithm that identifies a set of characteristics points where the statistics or distributions of the trajectory changes rapidly. The second step generates pseudo labels of the unlabelled segments by applying a clustering method: this requires a small proportion of trajectory segments to be labelled as input. The last step trains a classification model for activity recognition utilising the segments and their pseudo labels.

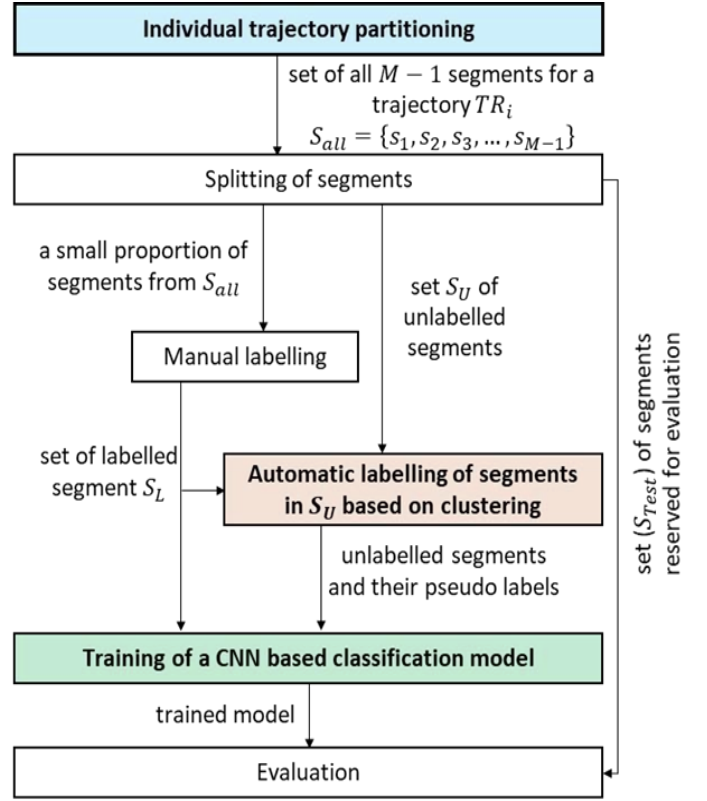

**Fig. 7.** A simplified schematic diagram of the proposed approach for activity recognition.

## A. Trajectory Partitioning

We aim to partition the individual trajectories into non-overlapping segments. The key idea is to identify a set of $M$ characteristics points (or change points) $CP = \{(p_{c_1}, t_{c_1}), (p_{c_2}, t_{c_2}), ..., (p_{c_M}, t_{c_M})\}$: $t_{c_1} < t_{c_2} < \cdots < t_{c_M}$ from each trajectory $TR_i \in \mathcal{T}$ and use those to partition each trajectory into $M - 1$ segments. These segments represent the different movement patterns in an individual trajectory.

To discover the characteristics points from individual trajectories, we apply an information theoretic criterion, Minimum Description Length (MDL) [55]. There are two desirable properties of the trajectory partitioning [56]: *preciseness* and *conciseness*. Preciseness represents the accuracy in which a set of chosen trajectory segments represent the original trajectory, whereas conciseness represents the number of segments (i.e., model parameters) used within its representation. These two properties are contradictory. For example, the preciseness is maximised (and conciseness is minimised) if we consider all the points within a trajectory as characteristics points. Likewise, the conciseness is maximised (and preciseness is minimised) if only the two end points of the trajectory are considered as the characteristic points. The MDL principle identifies the characteristics points of the trajectories by finding the optimal trade-off between the preciseness and conciseness properties [11, 56].

The function defining the MDL principle is given in (1) where $H$ represents a hypothesis, $D$ is the data, $L(H)$ is the description length of the hypothesis and $L(D|H)$ is the description length of the data encoded using the hypothesis, both expressed in bits. The hypothesis $H$ with the minimum $MDL$ is the one that achieves the highest data compression, or equivalently, the best explanation of the data. For our segmentation task, the hypothesis $H$ corresponds to the set of partitions of our trajectory data. Therefore, finding the optimal partitioning of the trajectories can be translated into finding the best hypothesis based on MDL.

$$MDL = L(H) + L(D|H) \qquad (1)$$

The two terms of the MDL function can be formulated using (2) and (3), respectively [56]. $L(H)$ in (2) represents the total length of all trajectory partitions where $len\left(p_{c_j}p_{c_{j+1}}\right)$ is the length of a line segment $\left(p_{c_j}p_{c_{j+1}}\right)$ that is computed using the Euclidean distance between two consecutive characteristics points $p_{c_j}$ and $p_{c_{j+1}}$. On the other hand, the formulation of $L(D|H)$ in (3) refers to the sum of the difference between a trajectory and a set of its trajectory partitions. For each partition, the difference between partition and the representing line segment is computed by taking the sum of the perpendicular distance $d_\perp\left(p_{c_j}p_{c_{j+1}}, p_k p_{k+1}\right)$ and the angular distance $d_\theta\left(p_{c_j}p_{c_{j+1}}, p_k p_{k+1}\right)$.

$$L(H) = \sum_{j=1}^{M-1} log_2\left(len\left(p_{c_j}p_{c_{j+1}}\right)\right) \qquad (2)$$

$$L(D|H) = \sum_{j=1}^{M-1}\sum_{k=c_j}^{c_{j+1}-1}\left\{ log_2\left(d_\perp\left(p_{c_j}p_{c_{j+1}}, p_k p_{k+1}\right)\right) + log_2\left(d_\theta\left(p_{c_j}p_{c_{j+1}}, p_k p_{k+1}\right)\right)\right\} \qquad (3)$$

The above formulation of $L(H)$ represents a measure of the conciseness, where as $L(D|H)$ indicates a measure of the preciseness. For our segmentation task, $L(H)$ increases as the number to partitions increases. On the other hand, larger deviations between the set of trajectory partitions and original trajectory causes $L(D|H)$ to increase. We aim to find the optimal partitioning that minimises the sum of $L(H)$ and $L(D|H)$.

Moreover, the cost of finding the optimal partitioning for a trajectory is exhaustive as it requires every subset of points in the trajectory to be considered. Hence, an approximate algorithm [56] with time complexity of $O(N)$ has been applied which considers a set of local optima as the global optimum. Let $MDL_{par}(p_i p_j)$ represents the MDL cost (i.e., $L(H) + L(D|H)$) of a trajectory between two points $p_i$ and $p_j$ ($i < j$) considering $p_i$ and $p_j$ are the only characteristic points, whereas $MDL_{nonpar}(p_i p_j)$ is the MDL cost when persevering the original trajectory – i.e., if there is no characteristic point between $p_i$ and $p_j$. It is obvious that $L(D|H)$ in $MDL_{nonpar}(p_i p_j)$ is zero. Hence, a local optimum is the longest trajectory partition $p_i p_j$ which satisfies the condition $MDL_{par}(p_i p_k) \leq MDL_{nonpar}(p_i p_k) : \forall_k \ i < k \leq j$. This means if $MDL_{par}(p_i p_k)$ smaller than $MDL_{nonpar}(p_i p_k)$, the selection of $p_k$ as a characteristic point will cause the MDL cost smaller compared to the MDL cost if $p_k$ is not chosen. The approximation process considers the first data point $p_1$ from the trajectory as the starting characteristic point and repeatedly compute $MDL_{par}$ and $MDL_{nonpar}$ for each subsequent point. In each step, if the cost of partitioning is equal or less than the cost of not partitioning, we increase the length of trajectory partition and continue computing the two costs for next point. Otherwise, we consider the previous point as the characteristics point and repeat the same procedure to search the next characteristic point until all data points are checked.

### B. Cluster and Label

In this section, we introduce the method for labelling the trajectory segments generated by the partitioning process. The main idea here is to generate a set of pseudo labels for the trajectory segments by applying constrained agglomerative hierarchical clustering with a small proportion of labelled segments.

Let $S_L = \left\{ (s_{l_1}, y_{l_1}), (s_{l_2}, y_{l_2}), \dots, (s_i, y_{l_i}) \right\}$ and $S_U = \left\{ s_{u_1}, s_{u_2}, \dots, s_{u_j} \right\}$ are the sets of labelled and unlabelled segments, respectively. The members of segments in $S_L$ are called seeds. Our task is to automatically label the trajectory segments in $S_U$ so that we can train a classification model using the data samples (e.g., segments in both $S_L$ and $S_U$). The classification model will then be applied to predict the class labels of the data samples in $S_{Test}$ that represent the set of test segments.

We adopt the hierarchical clustering method to generates labels for trajectory segments in $S_U$. In contrast to traditional clustering approach, we introduce constraints to the clustering structure to define how the seeds should be grouped. Specifically, we impose *cannot-link* constraints on the elements of $S_L$ to ensure that no more than one seed should be present in any cluster, even if the seeds have the same label. Each of the clusters will be comprised of trajectory segments that are most similar to its seed, and hence, its members (i.e., segments from $S_U$) will be labelled according to the class of its seed.

When applying constraint agglomerative hierarchical clustering, it is very important to choose an appropriate distance metric for the linkage criteria used to compute the similarity between clusters. The standard metrics based on point-to-point distance are not suitable for spatial data especially when the length of the time series are not same, as in our case [57, 58]. Hence, we apply the Hausdorff distance to compute the distance between pairs of trajectory segments. Considering the segments as geometric curves, the Hausdorff distance from a set of points $A$ to another set of points $B$ is the maximum distance of a set $A$ to the nearest point in the set $B$ and is defined as in (5).

$$HD_{A \to B} = \max_{a \in A} \left\{ \min_{b \in B} d(a,b) \right\} \quad (5)$$

where $d(a,b)$ is the distance between points $a$ and $b$ computed using any chosen distance metric (e.g., Euclidean distance). Generally, the Hausdorff distance is directed, which means that distance $HD_{A \to B}$ from $A$ to $B$ is not equal to the distance $HD_{B \to A}$ from $B$ to $A$. An undirected Hausdorff distance can be computed by taking the average of the two directed distances as in (6).

$$HD_{A \to B} = HD_{B \to A} = mean(HD_{A \to B}, HD_{B \to A)} \quad (6)$$

The presented clustering method can be explained from a graph-theoretic perspective. Let $F = S_L \cup S_U$ be the set of all training observations. Consider $G = (V, E)$ as an undirected graph. The set of vertices $V$ in $G$ represents all the observations in $F$ and a super-vertex ($*$). On the other hand, $E$ represents the edges among the vertices for all elements in $F$ as well as the edges between the super-vertex and vertices representing all the seeds. The weights of the first group of edges corresponds to the distance of two observations (i.e., segments) computed using the undirected Hausdorff distance. On the other hand, for the edges between the super-vertex ($*$) and vertices representing the seeds, the weights are set to zero. The labelling of the observations in $S_U$ then can be viewed as a specific way of finding a minimum spanning tree of $G$ using Kruskal's algorithm considering that the forest which is successively joined by Kruskal's algorithm is a set of clusters generated by the agglomerative hierarchical clustering dendrogram [59]. Since the weights of the edges between the super-vertex and vertices representing the seeds (i.e., the labelled segments in $S_L$) is zero, Kruskal's algorithm will add all the seeds at the beginning to the minimum spanning tree in the first $R$ iterations where $R$ is the cardinality of $S_L$. This will leave $R$ branches in the tree – these branches are called the main branches. The tree will then grow along the main branches in the successive iterations without created any new branches since all the edges from the super-vertex have already been added by $R$ iterations. This is equivalent to imposing *cannot-link* constraints in agglomerative hierarchical clustering between each pair of seeds. Upon termination of the algorithm, each of the branches in the tree corresponds to a cluster in the agglomerative hierarchical clustering.

TABLE I. PARAMETERS OF THE CNNS USED FOR STEPWISE SEARCH

| Parameters | Description | Values used for grid search |
|---|---|---|
| Filters | number of convolutional filters or kernels. $(F_1, F_2, \dots F_L)$ indicates network consists of $L$ convolutional layers and each layer $l$ consist of $F_l$ filters | [(16), (32), (64), (128), (32, 16), (32, 32), (64, 32), (64, 16), (128, 64), (64, 32, 16), (128, 64, 32, 16)] |
| Kernel size | length of the convolution filters | [3, 5] |
| Activation | activation functions for convolutional layers | ['relu', 'sigmoid', 'tanh'] |
| Strides | distance between two successive kernel positions is called a stride | [1,2,3] |
| Padding | how the centre of each kernel to overlap the outermost element of the inputs | ['valid, 'same'] |
| Kernel initializer | how to set initials weights of the convolutional filters | ['glorot_uniform', random_uniform'] |
| Pooling | pooling operation to perform on convolutional feature map | [None, 'max', 'avg'] |
| Dropout rate | fraction of neurons and their associated weights to disregard at each training epoch | [0, 0.1, 0.2, 0.3] |
| Dense neurons | number of layers in fully connected layers. $(N_1, N_2, \dots N_L)$ indicates $L$ fully connected layers and each layer $l$ consist of $N_l$ neurons | [(4,), (10, 4), (20, 4)] |
| Dense activation | activation functions for neurons in fully connected layers | ['relu', and/or 'softmax'] |
| Batch size | number of training samples per gradient update | [64, 128] |
| Epochs | number of epochs to train the model | [250, 500] |

### C. Classification model

To develop the classification models, we apply CNNs. The rationale of selecting CNNs are their ability to automatically learn features that represent meaningful patterns from large spatial or temporal datasets.

The CNNs used in this paper consist of multiple convolutional and pooling layers. Hence, the hyper-parameters of the networks have a significant influence on generalization ability, robustness, and overall predictive performance of the models. We find the optimal topology of CNNs and tune their hyper-parameters based on a stepwise search method. Specifically, we optimise one parameter at a time while keeping the remaining parameters unchanged by training the CNNs using training data and evaluating its performance on the validation data. We apply a stepwise search instead of an exhaustive grid search to reduce the training time. We consider CNNs up to 4 convolutional layers with a different number of filters in the range of 16 to 128, multiple filter sizes in each convolutional layer in the range of 3 to 5, two different types of pooling layers – max and average pooling, and multiple combinations of fully connected layers with a different number of nodes.

To optimise the parameters, we apply Adam optimization algorithm [60] with a differing number of epochs, minimizing the sparse categorical cross entropy, and applying dropout at each layer. Dropout [61] is a regularization method that randomly chooses specified fraction of nodes at each training epoch and disables their connection, hence disregards them during weight optimization. This technique has been found very effective for deep learning models to reduce over-fitting. TABLE I presents the entire search space considered to find the optimal structure and tune hyper-parameters of CNNs.

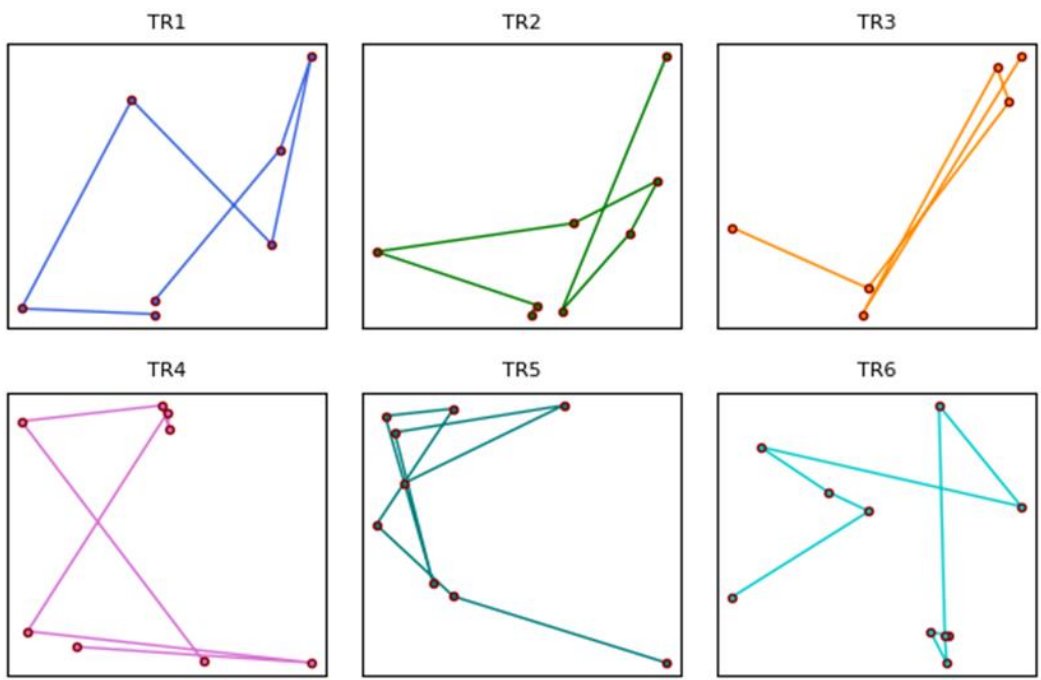

**Fig. 8.** Partitioning of trajectories. Lines indicate part (randomly selected) of trajectories and circles represent the identified partitioning points. The raw noisy data points between partitioning points are not shown for better representation.

The selected best architectures of the CNNs models for all the trajectories have similar structures with number of convolutional layers between 1 to 2. As an example, the best architectures obtained based on stepwise searching for the trajectory data for the operator at rig 6 includes 2 1D convolutional layers with 32 and 16 filters respectively. Both convolutional layers share the same kernel size, strides, padding method, and activation function: 3, 1, 'same', and 'tanh', respectively. Moreover, each convolutional layer is also followed by a max pooling layer, a dropout layer with dropout fraction of 0.2. The fully connected component consists of one dense layer with a *softmax* activation function and provides a probability of the 4 classes of activities. The weights of the CNNs were optimised with a maximum of 500 epochs and batch size of 128.

For each trajectory in our dataset, we develop a separate classification model using CNNs. Inputs to the CNNs models include the trajectory segments along with duration of the segments. Each of the models has been trained using the observations in both in $S_L$ and $S_U = \left\{ \left(s_{u_1}, y_{pl_1}\right), \left(s_{u_2}, y_{pl_2}\right), \dots, \left(s_{u_j}, y_{pl_j}\right) \right\}$ where $s_{u_{i\,(1 \leq i \leq j)}}$ is the an element of $S_U$ and $y_{pl_i}$ the pseudo label for $s_{u_i}$. This means the training data for each trajectory contains both the segments in $S_L$ with their actual class labels, and the segments in $S_U$ with the pseudo labels generated in clustering step.

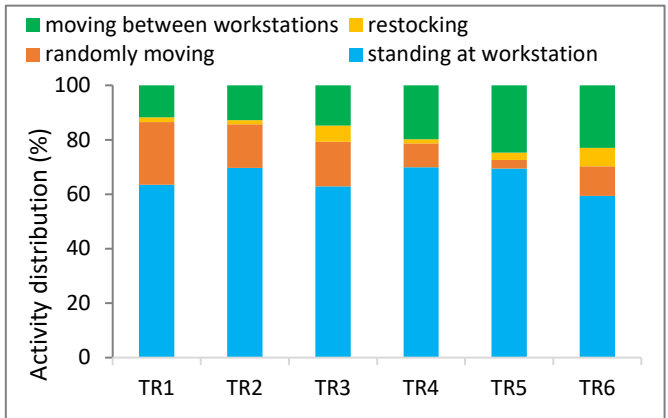

**Fig. 9.** Distribution of activities of workers into 4 defined categories.

## VI. EXPERIMENTS AND RESULTS

The partitioning algorithm is applied to each trajectory $TR_i$ in $\mathcal{T} = \{TR_i\}_{i=1}^{W}$ separately. **Fig. 8** shows the partitioning results (the partition points and the segments joining those points) for a small proportion of data points (2-3 minutes) from each trajectory. The raw noisy trajectory data points are not shown for better visualisation. As we can see, the MDL based algorithm correctly identify the trajectory segments with different characteristics – the portioning points are chosen where the behaviour of the trajectories changes significantly. The segments have different lengths and represent different activities of the workers in the factory. TABLE II presents the summary of trajectory partitioning results. The number of partitioning points for the trajectories varies in the range of 223 for $TR_1$ to 388 for $TR_6$. The variation in the number of partitioning points is expected since different workers show different movement patterns. **Fig. 9** presents the distribution of activities of the workers into predefined categories: *standing at workstation*, *moving randomly*, *moving between workstations*, *restocking*. More than 50% of segments of all trajectories are labelled as *standing at workstation* which highlights that the workers are mostly busy completing their assigned task for a majority of the time. The workers also spend a significant proportion of time moving between different workstations – the proportion of segments classified as *moving between workstations* varies between 11.71% to 24.71%. This is due to the collaborative nature of the tricycle assembly task and can be explained by the workflow in **Fig. 1**. As shown in **Fig. 1**, during the assembly process, the components built at each workstation needs to be delivered to another co-worker until the completion of work in progress.

TABLE II. SUMMARY INFORMATION ON TRAJECTORY SEGMENTATION

| Trajectory | Duration (sec) | Length | Partition points | Total segments |
|---|---|---|---|---|
| $TR_1$ | 3790 | 7727 | 223 | 222 |
| $TR_2$ | 5214 | 7840 | 328 | 327 |
| $TR_3$ | 3475 | 7266 | 252 | 251 |
| $TR_4$ | 3107 | 7712 | 264 | 263 |
| $TR_5$ | 7194 | 7803 | 341 | 340 |
| $TR_6$ | 7356 | 7847 | 388 | 387 |

Moreover, *moving randomly* is the third most frequent observed activity. The worker at rig 1 has the highest percentage of segments labelled as *moving randomly* (22.97% segments) followed by the worker at rig 3 (16.33%) and rig 2 (15.90% segments). The percentage of segments labelled as *moving randomly* is relatively low for the workers at the other three rigs. The high percentage of *moving randomly* segments for workers of the rig 1, 2, and 3 is due to the fact that they start working earlier compared to the workers at rigs 4, 5, and 6. This is because workers at the later rigs require their components to be built by the other rigs before they can commence their tasks. This allows the workers at rig 1, 2, and 4 to complete their assign tasks earlier and visit outside assembly area for taking break, meeting colleagues, etc. The *restocking* is the least frequent activity shown by all workers, 1.53% to 6.73% segments are labelled as *restocking*. This is expected since all the workstations hold the components they require to assemble 3 tricycles. Workers are then required to occasionally restock the parts until 6 tricycles are completed within 3 hours.

For each trajectory $TR_i$ in $\mathcal{T} = \{TR_i\}_{i=1}^{W}$ we aim to develop a separate semi-supervised classification model. To achieve this, the spatial trajectory segments for each trajectory $TR_i$ are divided into 3 non-overlapping subsets: $S_L$, $S_U$, and $S_{Test}$. $S_L$ contains only very small proportion (20%) of trajectory segments that are manually labelled. $S_U$ and $S_{Test}$ respectively contains 80% and 20% of the remaining segments. The segments in $S_U$ are labelled using the constrained agglomerative hierarchical clustering. The segments in both $S_L$ and $S_U$ are used to train the classification model whereas the segments in $S_{Test}$ are used to evaluate the accuracy of the classification model.

To evaluate the performance of the models we use two different metrics: misclassification rate and F-score. Misclassification Rate (MCR) refers to the percentage of observations that were incorrectly predicted by the classification model. MCR has a range of 0 to 1: a lower value of MCR indicates a higher accuracy. The F-score is a standard measure for classification and is defined as in (7) for a binary classification problem. For multi-class classification task as ours, this is the average of the F-score of each class with weighting determined by the number of observations in each class. The F-score also has a range of 0 to 1, with 1 being the most accurate classifier and 0 indicating the worst possible classifier. Although we present the numerical results using both metrics, F-score will be considered as the primary metric for analysis of prediction accuracy.

$$Fscore = 2 \times \frac{(precision \times recall)}{(precision + recall)} \quad (7)$$

where *precision* is the fraction of observations that the model classified as positive that were true positives and *recall* refers to the fraction of true positive observations that the model correctly classified as positive.

TABLE III. ACCURACY OF THE SEMI-SUPERVISED APPROACH IN TERMS OF THE F-SCORE AND MISCLASSIFICATION RATE

| Trajectory | F-score | MCR |
|---|---|---|
| $TR_1$ | 0.81 | 0.19 |
| $TR_2$ | 0.92 | 0.08 |
| $TR_3$ | 0.85 | 0.15 |
| $TR_4$ | 0.93 | 0.07 |
| $TR_5$ | 0.95 | 0.05 |
| $TR_6$ | 0.85 | 0.15 |

TABLE III presents the classification accuracy evaluated on the trajectory segments in $S_{Test}$ for each trajectory $TR_i$ in $\mathcal{T} = \{TR_i\}_{i=1}^{W}$. The classification accuracy (in terms of F-score) for the trajectories generated by the workers at all rigs is 81% or higher. The higher classification accuracy indicates the ability of the proposed approach to classify the activities of the workers with using only a small set of labelled segments. Amongst the trajectories of all six workers, the trajectory for the worker at rig 5 achieved the highest classification accuracy followed by the trajectory for workers at rig 4 and 2, respectively – their F-scores are above 90%. On the other hand, the classification accuracy was lowest for the trajectory of the worker at rig 1. The F-scores of the trajectories of two other workers is 85% The main reason for relatively lower accuracy for the trajectory of worker at rig 1 is the smaller number of training samples used in $S_L$. There are 44 segments in $S_L$ for trajectories of the worker at rig 1 compared 65 segments in $S_L$ for trajectories of the worker at rig 2 for example. The higher number of labelled segments in $S_L$ help to generate pseudo labels with better accuracy confidence. For example, the accuracy of the pseudo labels generated by the hierarchical clustering is 80% and 90% for trajectories for workers at rig 1 and 2, respectively. Using the more accurate pseudo labels helps to train the final CNN models to better learn the patterns which consequently reflected in the final accuracy results.

Moreover, we reserve 20% of labelled segments into $S_L$ for each trajectory as previously mentioned. These segments (called seeds) are then used in the clustering phase to generate pseudo labels for the segments in $S_U$. Hence, it is important to check the influence this selection of labelled segments on the performance of the semi-supervised classification model. We repeated all the experiments with different proportions of labelled data in $S_L$: 5% to 20% with 5% increments. **Fig. 10** presents the F-scores of the classification models with respect to different proportions of segments in $S_L$ for each trajectory. The classification accuracy of all trajectories monotonically increases as the proportion of labelled segments in $S_L$ increase from 5% to 20% The improvement in classification accuracy is expected since a higher proportion of segments in $S_L$ provides more information for the clustering algorithm to exploit for pseudo label generation. In other words, more manually labelled segments means availability of more patterns of different activity types during clustering algorithm which consequently helps to the clustering process to generate more accurate the pseudo labels for segments in $S_U$. These results provide more accurate training of the final prediction models.

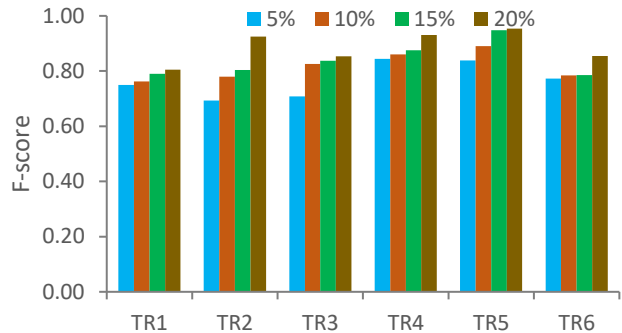

**Fig. 10.** Classification accuracy (F-score) of the model with different proportion of manually labelled segments in $S_L$ used to generate pseudo labels for segment in $S_U$.

The predicted labels of activities for the trajectory segments can be used in diverse ways for decision making purposes. For example, the distribution of activities of workers during their shift (as shown in **Fig. 9**) can help to improve worker productivity. As we can see, the worker at rig 1 spends second highest proportion of time doing random movements. The worker at rig 1 is the first person in the workflow and likely to complete the assigned task before any of the other workers. Moreover, from the data it is evident that worker at rig 1 makes most of the random moves later part of shift. This time can be better utilised doing productive work as decided by the management. Similarly, the worker at rig 5 spends second highest time on moving between workstations as per distribution of activities in **Fig. 9**. If the movements are happening to search for missing components or tools, those can be made available at rig 5 to reduce the searching time.

TABLE IV. CENTROIDS OF THE TEN CLUSTERS REPRESENTING JOINT ACTIVITIES OF WORKERS. WE USE SHORT NAMES FOR FOUR ACTIVITY CLASSES (STAND: *STANDING AT WORKSTATION*, MOVE: *MOVE BETWEEN WORKSTATIONS*, RANDOM: *RANDOMLY MOVING*, AND RESTOCK: *RESTOCKING*.)

| Cluster ID | Centroid [Concurrent activities of 6 workers] |
|---|---|
| 1 | ['stand', 'stand', 'stand', 'stand', 'stand', 'stand'] |
| 2 | ['stand', 'stand', 'stand', 'move', 'stand', 'stand'] |
| 3 | ['stand', 'stand', 'stand', 'stand', 'stand', 'move'] |
| 4 | ['stand', 'stand', 'move', 'stand', 'stand', 'stand'] |
| 5 | ['stand', 'stand', 'move', 'random', 'stand', 'stand'] |
| 6 | ['stand', 'random', 'random', 'stand', 'stand', 'stand'] |
| 7 | [restock, 'stand', 'stand', 'stand', 'stand', 'stand'] |
| 8 | ['random', 'stand', 'stand', 'stand', 'stand', 'stand'] |
| 9 | ['move', 'stand', 'stand', 'stand', 'stand', 'stand'] |
| 10 | ['stand', 'stand', 'stand', 'stand', 'move', 'stand'] |

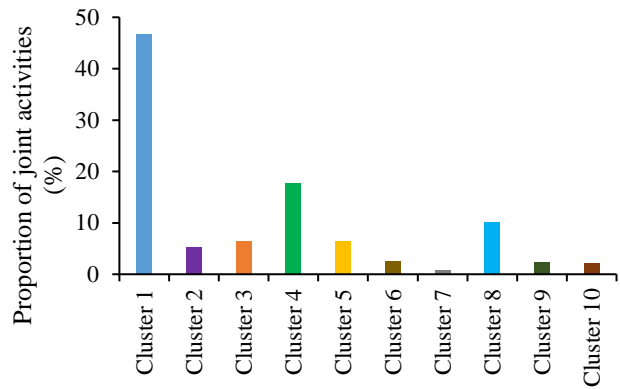

**Fig. 11.** Cluster (State) distribution of the joint activity data

Another interesting inference that can be made from the activity classification results is 'factory floor state'. A state can be defined in numerous ways. In this case, we refer to the joint behaviour of workers as a state, i.e., what the workers are doing concurrently. We clustered the joint activity class labels of all six workers to obtain the factory floor state. TABLE IV presents the centroids of the ten clusters obtained using K-mode algorithm [62]. **Fig. 11** shows the distribution of clusters over joint activity data. Workers working at their respective stations is the most common state (cluster 1) and it constitutes 46.6% of the joint behaviour classes. The second highest cluster observed is cluster 4 where most workers are at their rigs except worker at rig 3 who is moving between stations. We also observe a lot of movement of that worker from rig 3 to rig 2 and this observation aligns well with the workflow (see **Fig. 1**). The third most common state is when most workers are at their rigs except the worker at rig 1 who is moving randomly (cluster 8). As discussed previously, the worker at rig 1 starts the assigned task at the beginning of the workflow with no dependencies hence finishes work early and moves randomly later while other workers are at their rigs. Such state estimation through clustering can summarise concisely what's happening at the factory floor and can be used for operational decision making.

## VII. CONCLUSIONS

We presented an approach for human activity recognition from noisy indoor trajectory data and studied its application in manufacturing context. The proposed approach adopted the concept of semi-supervised learning: generated pseudo labels based on constraint hierarchical clustering and trained convolutional neural networks as the classifiers that used the trajectory segments as inputs and respective pseudo labels as outputs. The approach is comprehensively evaluated using six trajectories of human workers at a tricycle assembly workshop. Results indicate that proposed approach can accurately classify the activities of the workers at different part of their trajectories – the classification accuracy in terms of F-score varies between 0.81 to 0.95. Moreover, this performance is achieved with small proportion of labelled examples. The key advantage of this approach over existing supervised activity recognition models is that it saves time and resources required for manual labelling of input-output examples by the domain experts. In addition, although the approach is developed to identify four target activities (*standing at workstation*, *moving randomly*, *moving between workstations*, *restocking*) specific to manufacturing environment, it is generic and can be applied to such other activities as well. Future research will focus on applying the approach on the trajectory datasets from other domains and improving pseudo label generation process based on advanced self-supervised methods. Another avenue of future work is representing the segments using raster images overlayed on factory layout and then classify them.